\documentclass[referee, sn-mathphys-num]{sn-jnl}

\usepackage{graphicx}%
\usepackage{multirow}%
\usepackage{amsmath,amssymb,amsfonts}%
\usepackage{amsthm}%
\usepackage{mathrsfs}%
\usepackage[title]{appendix}%
\usepackage{xcolor}%
\usepackage{textcomp}%
\usepackage{manyfoot}%
\usepackage{booktabs}%
\usepackage{algorithm}%
\usepackage{algorithmicx}%
\usepackage{algpseudocode}%
\usepackage{listings}%
\usepackage[version=4]{mhchem}
\usepackage{subcaption}
\usepackage{lineno}

\usepackage{longtable}
\theoremstyle{thmstyleone}%
\theoremstyle{thmstyletwo}%
\theoremstyle{thmstylethree}%
\begin{document}

\title{Deep Operator Networks for Surrogate Modeling of Cyclic Adsorption Processes with Varying Initial Conditions}

\author[1]{\fnm{Beatrice} \sur{Ceccanti}}\email{beatrice.ceccanti@gmail.com}
\equalcont{These authors contributed equally to this work.}

\author[1]{\fnm{Mattia} \sur{Galanti}}\email{m.galanti@tue.nl}
\equalcont{These authors contributed equally to this work.}

\author[1]{\fnm{Ivo} \sur{Roghair}}\email{I.Roghair@tue.nl}

\author*[1]{\fnm{Martin} \sur{van Sint Annaland}}\email{M.v.SintAnnaland@tue.nl}

\affil*[1]{\orgdiv{Department of Chemical Engineering and Chemistry}, \orgname{Eindhoven University of Technology}, \orgaddress{\street{Het Kranenveld 14}, \city{Eindhoven}, \postcode{5612AV}, \country{The Netherlands}}}

\abstract{Deep Operator Networks are emerging as fundamental tools among various neural network types to learn mappings between function spaces, and have recently gained attention due to their ability to approximate nonlinear operators. In particular, DeepONets offer a natural formulation for PDE solving, since the solution of a partial differential equation can be interpreted as an operator mapping an initial condition to its corresponding solution field. In this work, we applied DeepONets in the context of process modeling for adsorption technologies, to assess their feasibility as surrogates for cyclic adsorption process simulation and optimization. The goal is to accelerate convergence of cyclic processes such as Temperature-Vacuum Swing Adsorption (TVSA), which require repeated solution of transient PDEs, which are computationally expensive. Since each step of a cyclic adsorption process starts from the final state of the preceding step, effective surrogate modeling requires generalization across a wide range of initial conditions. The governing equations exhibit steep traveling fronts, providing a demanding benchmark for operator learning. To evaluate functional generalization under these conditions, we construct a mixed training dataset composed of heterogeneous initial conditions and train DeepONets to approximate the corresponding solution operators. The trained models are then tested on initial conditions outside the parameter ranges used during training, as well as on completely unseen functional forms. The results demonstrate accurate predictions both within and beyond the training distribution, highlighting DeepONets as potential efficient surrogates for accelerating cyclic adsorption simulations and optimization workflows.}




\maketitle
\section{Introduction}\label{sec:introduction}
The rapid proliferation of artificial intelligence across scientific disciplines has positioned machine learning as a central tool for data-driven modeling and prediction \cite{Duede2024OilFields, Pires2023ArtificialArt}. Within this landscape, neural-network-based methods have become particularly prominent due to their expressive power in approximating highly nonlinear mappings \cite{Goodfellow2016DeepLearning}. Beyond classical architectures such as feedforward, convolutional, and recurrent neural networks, which are typically designed to learn mappings between finite-dimensional input and output spaces, recent research has increasingly focused on models that approximate mappings between infinite-dimensional function spaces. In this operator-learning setting, the objective is to learn a nonlinear operator that maps an input function to an output function.  

Deep Operator Networks (DeepONets) were introduced in this context as a principled framework for approximating such nonlinear operators, making them especially well-suited for learning solution operators of partial differential equations and other physics-based systems \cite{Lu2021LearningOperators}. In the original work by Lu et al., DeepONet was tested on several problems and families of PDEs to establish its capability to approximate nonlinear solution operators under controlled settings. 

Following its introduction in 2021, a growing body of literature has explored the use of DeepONets for solving PDEs across different physical domains, often proposing task-specific adaptations or extensions of the original architecture. However, in most of these studies, the primary emphasis lies on generalization with respect to governing equations, parametric variations, or geometrical configurations. 

By contrast, initial conditions are typically treated in a more abstract manner, commonly represented as samples drawn from Gaussian Processes or related stochastic function generators, whose purpose is to span the input function space rather than to reflect physically realizable system states \cite{Kumar2025SynergisticSolving, Herde2024Poseidon:PDEs, Wang2021LearningDeepONets}. While such formulations are well-suited for benchmarking operator-learning performance, they do not explicitly address scenarios in which the structure and physical consistency of the initial condition play a central role in the system dynamics.

This limitation becomes particularly relevant in cyclic and multi-stage processes, in which the system dynamics are governed by a sequence of coupled transient steps rather than by a single forward evolution. A representative example is cyclic adsorption-based separation processes, such as temperature swing adsorption (TSA), vacuum swing adsorption (VSA), or temperature–vacuum swing adsorption (TVSA), which are widely used in gas separation and carbon capture applications \cite{DouglasLeVan2019AdsorptionExchange, Duong1998AdsorptionKinetics}. In these processes, the operating cycle is composed of successive steps (including adsorption, regeneration, and conditioning phases), each of which starts from the final state reached at the end of the preceding step. As a result, the initial condition of every step is not arbitrary but is instead constrained by the state variables inherited from the previous stage of the cycle. 

Moreover, from a computational perspective, the simulation of cyclic adsorption processes is particularly demanding. Unlike single-pass transient problems, these systems do not admit a steady state in the classical sense. Instead, they converge to a cyclic steady state (CSS), defined as a periodic regime in which the system state at the beginning of a cycle, consisting of several sub-steps with different inlet conditions, coincides with the state at the end of the previous one. The CSS is therefore not known a priori and must be identified iteratively by repeatedly solving a coupled system of transient partial differential equations until convergence is achieved between the final state of one cycle and the initial state of the next. The governing equations typically include mass, momentum, and energy balances and involve a large number of operating and material parameters, such as volumetric flow rates, mass, momentum, heat transfer coefficients, and sorbent thermophysical properties.

Exploring the system response across different physically admissible initial states and operating conditions therefore requires the repeated numerical solution of high-dimensional PDE systems. As a consequence, high-fidelity simulations become prohibitively expensive when embedded within multi-criteria optimization, design-space exploration, or real-time control frameworks. To mitigate this computational bottleneck, recent research has increasingly focused on surrogate modeling approaches based on neural networks \cite{Galanti2025TowardsNetworks, Subraveti2024ASimulations, Ye2019ArtificialAdsorption, Leperi2019110thCapture}. In principle, such models enable rapid inference of the system dynamics, allowing efficient exploration of the solution landscape of cyclic processes. 

Therefore, in this work we investigate the use of Deep Operator Networks as surrogate models for cyclic adsorption processes, with a specific focus on their ability to generalize across physically meaningful initial conditions. Rather than addressing a generic benchmark settings, we consider a representative adsorption step extracted from a cyclic process, in which the initial state reflects the outcome of a previous cycle stage. To this end, we formulate a simplified yet physically consistent adsorption model and train DeepONets to predict the resulting spatiotemporal fields. Although physically motivated, these PDEs also provide a challenging test for neural operator learning due to steep traveling fronts and and strong sensitivity to initial conditions.

This paper is organized as follows. In Section~\ref{sec:methods}, the adopted methodology is presented, including the problem setup, data generation procedures, the DeepONet architecture, and the training strategy. Section~\ref{sec:results} reports the training outcomes, as well as the results obtained. Finally, Section~\ref{sec:discussion} discusses the main strengths and limitations of the proposed approach.

\section{Methods}\label{sec:methods}
\subsection{Problem Setup}\label{subsec:problem_setup}
We consider a nonlinear PDE system describing the dynamics of mass transport of a component through a packed bed filled with a solid adsorbent onto which the component is adsorbed, under the following hypotheses:
\begin{itemize}
    \item The system is modeled as one-dimensional along the axial direction of the adsorption column (radial gradients of concentration, temperature, or velocity are neglected). 
    \item Local thermodynamic equilibrium between gas and solid phase is assumed at the particle surface.
    \item The process is assumed to be isothermal. 
    \item  External mass transfer resistance at the film of the gas-solid interphase is the limiting mechanism among the mass transfer resistances. 
    \item Axial dispersion effects are assumed to be negligible under the conditions of the simulations and within the scope of the analysis presented in this work. 
    \item Only one species is present, that adsorbs according to a linear (Henry) adsorption isotherm. 
\end{itemize}

Under these assumptions, the governing equations can be written as: 
\begin{equation}
    \frac{\partial C_g}{\partial t} + v_x \frac{\partial C_g}{\partial x} =
- \frac{k_g a_s}{\varepsilon_B}
\left( C_g - \frac{C_s^{R}}{K_{eq}} \right)
\end{equation}
\begin{equation}
\frac{\partial C_s}{\partial t} = \frac{k_g a_s}{1 - \varepsilon_B}
\left(C_g - \frac{C_s^{R}}{K_{eq}}\right)
\end{equation}

where $C_g$ is the molar concentration in the gas phase, $C_s$ is the 
molar concentration adsorbed per unit volume of the solid adsorbent, 
and $C_s^{R}$ is the adsorbed-phase concentration at the external surface of 
the solid particles. Under the local equilibrium assumption, the concentration at the external surface  of the adsorbent is equal to the bulk solid concentration, which implies
\[
C_s^{R} = C_s.
\]
To preserve the generality of the formulation and facilitate comparisons across different physical scales, the governing equations are first expressed in terms of dimensionless space–time variables (assuming $v_x > 0$). Specifically, a change of variables is introduced to define the dimensionless coordinates $\tau$ and $\xi$ according to Eq.~\eqref{eq:tau_xi_scaling}. This transformation  recasts the problem in a non-dimensional form that is independent of the absolute spatial and temporal scales.

Subsequently, both dependent and independent variables are normalized with respect to characteristic reference values, as defined in Eq.~\eqref{eq:normalization}. By ensuring that all inputs and outputs remain within comparable and bounded ranges, the normalization improves numerical conditioning and enhances training stability, while preserving the physical structure of the governing equations. In particular, such scaling reduces the risk of vanishing or exploding gradients, which commonly arise in deep learning models when variables span multiple orders of magnitude \cite{Yu2023NormalizationNetworks}.

\begin{equation} \label{eq:tau_xi_scaling}
\tau = \frac{k_g a_s}{(1 - \varepsilon_B) K_{eq}}
 t ,
\qquad
\xi = \frac{k_g a_s}{\varepsilon_B v_x}\, x
\end{equation}

\begin{equation} \label{eq:normalization}
C_g^* = \frac{C_g}{C_0},
\qquad
C_s^* = \frac{C_s}{K_{eq} C_0},
\qquad
\tau^* = \frac{\tau}{\tau_0},
\qquad
\xi^* = \frac{\xi}{\xi_0}
\end{equation}

The final PDE system results in the following equations:
\begin{equation} \label{eq:gas_gov_eq}
\frac{\varepsilon_B}{(1-\varepsilon_B) K_{eq} \tau_0} 
\frac{\partial C_g^*}{\partial \tau^*} + 
\frac{1}{\xi_0} \frac{\partial C_g^*}{\partial \xi^*}
=  - \left( C_g^* - C_s^* \right)
\end{equation}

\begin{equation} \label{eq:solid_gov_eq}
\frac{1}{\tau_0} 
\frac{\partial C_s^*}{\partial \tau^*} = \left( C_g^* - C_s^* \right)
\end{equation}

with initial and boundary conditions reported below:
\begin{equation} \label{eq:Bcs}
C_g^* \big|_{(\xi^* = 0, \tau^*)} = 1,
\qquad
\left. \frac{\partial C_g^*}{\partial \xi^*} \right|_{(\xi^* = 1, \tau^*)} = 0
\end{equation}

\begin{equation} \label{eq:Ics}
C_g^* \big|_{(\xi^*, \tau^*=0)} = C_g^{*0}(\xi^*),
\qquad
C_s^* \big|_{(\xi^*, \tau^*=0)} = C_s^{*0}(\xi^*)
\end{equation}

The governing equations \eqref{eq:gas_gov_eq}–\eqref{eq:solid_gov_eq}, together with the boundary conditions \eqref{eq:Bcs}, define a nonlinear mapping from an initial concentration profile to the corresponding spatiotemporal solution on
\((\xi^*,\tau^*)\in(0,1)\times(0,1)\).

The gas-phase initial condition is denoted by
\[
C_g^{*0}(\xi^*) := C_g^*(\xi^*,0)
\]
At the initial time \(\tau^*=0\), local thermodynamic equilibrium between the gas and solid phases is assumed.  
As a result, the solid-phase initial condition is not independent but is uniquely determined by the adsorption isotherm.
Under this constraint, the PDE system induces two solution operators parameterized by the gas-phase initial profile:
\[
\mathcal{G}_g:\ C_g^{*0}(\xi^*) \longmapsto C_g^*(\xi^*,\tau^*),
\qquad
\mathcal{G}_s:\ C_g^{*0}(\xi^*) \longmapsto C_s^*(\xi^*,\tau^*),
\]
Therefore, two Deep Operator Networks are trained to approximate these operators:
\[
\widehat{\mathcal{G}}_g \approx \mathcal{G}_g,
\qquad
\widehat{\mathcal{G}}_s \approx \mathcal{G}_s
\]

\subsection{Dataset generation}\label{subsec:data_generation}
We built a composite dataset of 10,000 initial-condition functions. The classes of functions chosen for this work cover diverse functional families that can occur in an adsorption-desorption cycle:

\begin{itemize}
    \item Fully regenerated bed: the gas and solid phase concentrations are constant and equal to zero.
    \item Quasi-steady desorption: the desorption rate is slow and uniform; therefore, nearly flat concentration profiles arise.
    \item Moderate mass transfer regime: small but not negligible axial gradients may lead to approximately linear profiles when mass transfer and dispersion terms balance linearly.
    \item Absence of desorption: the lack of a successful adsorption step reproduces the sigmoid or exponential-like functions that result from the adsorption step.
    \item Complete adsorption and partial desorption: in this case, increasing sigmoids or exponential-like concentration profiles can occur.
    \item Partial adsorption and desorption: the incomplete steps can cause the presence of high concentration peaks that resemble Gaussian-like functions.
\end{itemize}

To represent these regimes in a controlled and parametric manner, the initial conditions were generated from four families of analytical functions: straight lines \(f(x)=mx+q\), sigmoidal functions \(f(x)=\left(1+e^{-k(x-c)}\right)^{-1}\), exponential functions \(f(x)=e^{\alpha(x-\beta)}\), and Gaussian functions \(f(x)=\exp\!\left(-\tfrac{(x-\mu)^2}{2\sigma^2}\right)\). Each function family contributes 25\% of the total dataset, resulting in a uniformly balanced distribution across the four classes.

\begin{figure}[ht!]
\centering
\includegraphics[width=0.8\textwidth]{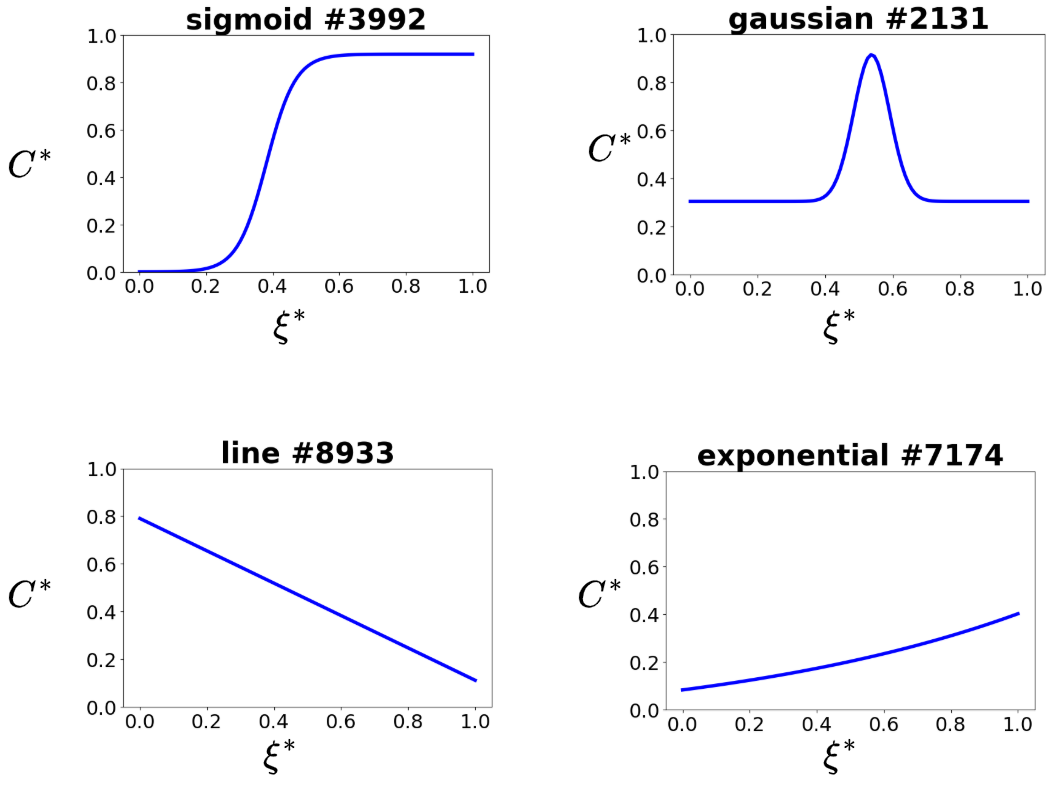}
\caption{Illustration of the input functions dataset, where $C^*$ represents either $C_g^*$ or $C_s^*$. Each title reports the functional type and the number of its position in the list of functions.}\label{fig:dataset_rescaled}
\end{figure}

The parameters of each function family were sampled randomly over prescribed ranges to ensure variability in amplitude, slope, curvature, and localization (see Table \ref{tab:ics_params}). Moreover, an additional random rescaling is applied such that the final function is also translated vertically, while still lying within the physically possible domain in \([0,1]\). The rescaled function is \(g(x) = a f(x) + b\), where \(a\) and \(b\) are the scaling factors. An example of some sampled rescaled initial conditions is shown in Fig. \ref{fig:dataset_rescaled}.

\begin{table}[h!] 
\centering
\caption{Ranges of the parameters of the functions used for dataset generation} \label{tab:ics_params}
\begin{tabular}{lcccc}
\hline
  \textbf{Gaussian} & \textbf{Sigmoid} & \textbf{Exponential} & \textbf{Line} \\
\hline
\( \mu \in [0,1] \) & 
\( k \in [5,30] \) & 
\( \alpha \in [-5,5] \) & 
\( m \in [-2,2] \) \\
\( \sigma \in [0.05,2] \) & 
\( c \in [0,1] \) & 
\( \beta = 0.5 \) & 
\( q \in [-1,1] \) \\
\hline
\end{tabular}
\end{table}

For each sampled initial condition, the governing equations
(Eqs.~\ref{eq:gas_gov_eq} and \ref{eq:solid_gov_eq}) were solved numerically using finite-difference schemes implemented in the \texttt{pymrm} package developed within our group \cite{Peters2025Pymrm:Modeling}. The spatial domain was discretized using 100 grid points, and the same spatial grid was used to represent both the initial condition functions and the resulting solution fields. Time integration was performed using 101 temporal points. It was verified that the selected number of grid cells and time steps provided sufficiently accurate numerical solutions. The set of simulation parameters used is reported in Table~\ref{tab:sims_pars}.

\begin{table}[h!]
\centering
\caption{Values and units of the parameters used}\label{tab:sims_pars}
\begin{tabular}{l l c c}
\toprule
\textbf{Definition} & \textbf{Description} & \textbf{Value} & \textbf{Unit} \\
\midrule
$L$        & Length of packed bed                         & 1     & m \\
$v_x$      & Superficial gas velocity                     & 0.1   & m\,s$^{-1}$ \\
$\varepsilon_B$ & Packed bed porosity                    & 0.5   & -- \\
$k_g$      & Mass transfer coefficient                    & 0.01  & m\,s$^{-1}$ \\
$d_p$      & Diameter of spherical particle               & 0.005 & m \\
$a_s$      & Specific surface area of a spherical particle& 1200  & m$^{-1}$ \\
$K_{eq}$   & Equilibrium constant                         & 100   & -- \\
$t_{tot}$  & Total simulation time                        & 1200  & s \\
$C_0$ & Reference initial condition concentration & 1000 & mol m$^{-3}$\\
\bottomrule
\end{tabular}
\end{table}

The full dataset of 10,000 initial conditions was randomly partitioned
into training, validation, and test subsets, as summarized in
Table~\ref{tab:ics_params_scaling}. The training set was used to optimize the network parameters via backpropagation, while the validation set was evaluated at each training epoch to monitor generalization performance and to select the model yielding the lowest validation error, thereby mitigating overfitting.

\begin{table}[h!] 
\centering
\caption{Information table of the generated datasets} \label{tab:ics_params_scaling}
\begin{tabular}{ll}
\hline
\textbf{Description} & \textbf{Value} \\
\hline
Initial condition matrix shape & 10,000 ICs $\times$ 100 space points \\
Each solution shape & 100 space points $\times$ 101 time points \\
Train set & 7200 samples \\
Val set & 1800 samples \\
Test set & 1000 samples \\
\hline
\end{tabular}
\end{table}

\subsubsection{Expanded Out-of-Distribution (OOD) Dataset}
To assess the robustness of the learned operator beyond the distribution
seen during training, we constructed an additional out-of-distribution
(OOD) dataset composed of 1000 initial-condition functions. This dataset probes two generalization modes: 
\begin{itemize}
    \item parameter extrapolation: obtained by extending the
parameter ranges of the function families used to generate the training
initial conditions.
    \item family shift: obtained by introducing a new functional family (sine waves) not included in the training set.
\end{itemize}

The sine family is defined as
\[
f(x)=\sin(w_0 x + \phi).
\]
In this dataset, each function family weights approximately 20\% of the total. Table~\ref{tab:ood_ranges} reports the parameter ranges used to generate the OOD initial conditions.  

\begin{table}[h!]
\centering
\caption{Parameter ranges used to generate the expanded (OOD) dataset}
\label{tab:ood_ranges}
\begin{tabular}{lccccc}
\hline
\textbf{Gaussian} & \textbf{Sigmoid} & \textbf{Exponential} & \textbf{Line} & \textbf{Sine} \\
\hline
$\mu \in [-0.5,\,1.5]$ &
$k \in [30,\,50]$ &
$\alpha \in [-7,\,7]$ &
$m \in [-4,-2)\,\lor\,[2,4)$ &
$w_0 \in [0.1,\,0.5]$ \\
$\sigma \in [0.2,\,0.4]$ &
$c \in [-0.8,\,1.2]$ &
$\beta \in [0.2,\,0.7]$ &
$q \in [-1,\,1]$ &
$\phi = 0$ \\
\hline
\end{tabular}
\end{table}

\subsection{DeepONets}\label{subsec:deeponets}
Supported by the universal approximation theorem \cite{Chen1995UniversalSystems}, Deep Operator Networks are neural networks designed to learn operators that are particularly useful for problems requiring a mapping from function to function.

In this work, DeepONets were implemented following the unstacked architecture introduced by Lu et al., which consists of a branch network and a trunk network \cite{Lu2020DeepONet:Operators}. The branch network encodes the initial condition functions evaluated at a set of spatial sensors (axial grid points), while the trunk network encodes the spatiotemporal coordinates \((\xi^*,\tau^*)\) at which the solution is queried. The same DeepONet architecture is used to approximate two solution operators, one for the gas-phase field and one for the solid-phase field. In both cases, the input to the branch network is the gas-phase initial concentration profile \(C_g^{*0}(\xi^*)\), while the outputs correspond to either the gas- or solid-phase spatiotemporal concentration field.

Using this architecture, each solution operator \(\mathcal{G}\) (gas or solid) is represented as a decomposition into a set of basis functions and corresponding coefficients. In particular, the trunk network learns a collection of spatiotemporal basis functions \(\{\boldsymbol{\phi}_k(\xi^*,\tau^*)\}_{k=1}^{p}\), while the branch network maps the gas-phase initial condition \(C_g^{*0}(\xi^*)\) to a set of coefficients \(\{a_k(C_g^{*0})\}_{k=1}^{p}\). The predicted solution fields are then reconstructed via an inner product between the branch and trunk representations, yielding the approximation
\[
\widehat{\mathcal{G}}\!\left(C_g^{*0}\right)(\xi^*,\tau^*)
=
\sum_{k=1}^{p}
a_k\!\left(C_g^{*0}\right)\,
\boldsymbol{\phi}_k(\xi^*,\tau^*).
\]

\begin{figure}[ht!]
\centering
\includegraphics[width=0.99\textwidth]{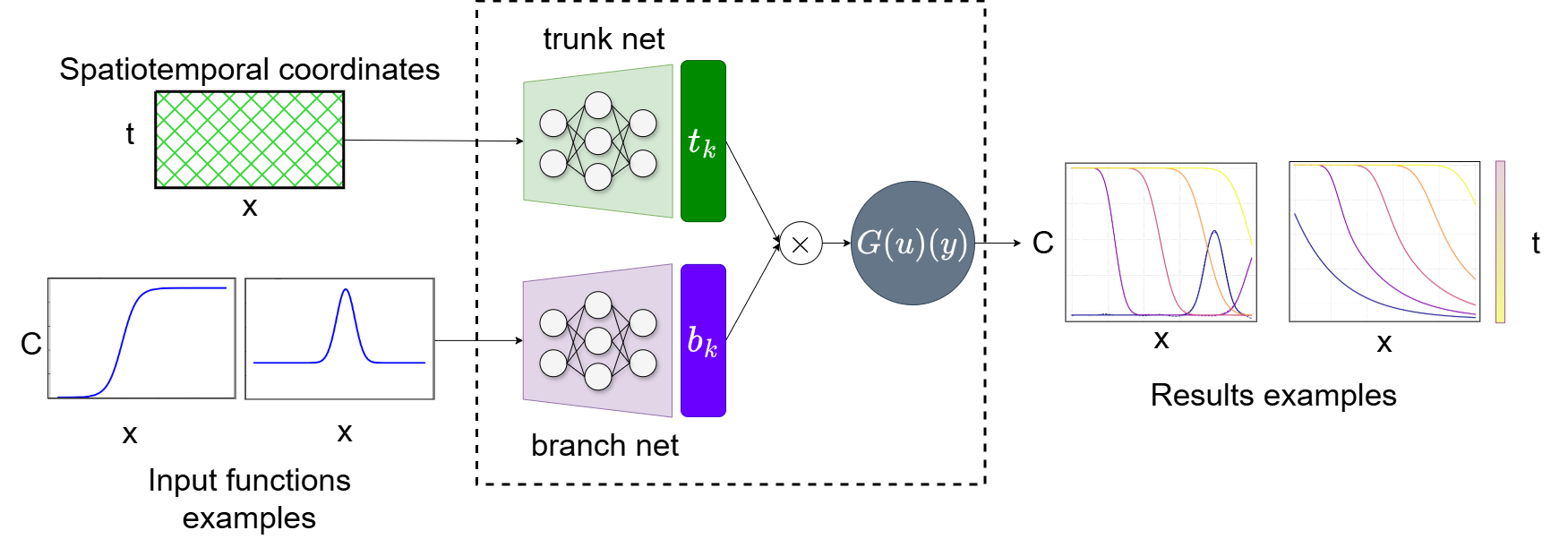}
\caption{Illustration of the DeepONet in the problem setup, in which the spatio-temporal coordinates as well as the input functions are fed to the trunk and the branch net, respectively. The network provides solutions in the spatio-temporal domain.}\label{fig:DeepONet_scheme}
\end{figure}

This formulation produces solution values evaluated at the same spatiotemporal grid provided to the trunk network, resulting in matrix-valued concentration fields consistent with the discretization used for the reference solutions. A schematic representation of the DeepONet architecture is shown in Fig.~\ref{fig:DeepONet_scheme}.

\subsubsection{Network architecture}
Both branch and trunk sub-networks are implemented as fully connected multilayer perceptrons with identical depth and width. The architecture specifications of the DeepONet have been summarized in Table \ref{tab:architecture}. The trunk and branch networks each consist of six hidden layers with 200 neurons per layer. The dimensionality of the latent space (i.e.\ the number of learned basis functions in the last layer) is set to $100$.

Different activation functions are used in the two sub-networks.
The branch network employs the SiLU activation function, which provides smooth nonlinear
transformations and has been found to be robust for encoding heterogeneous initial conditions.
The trunk network employs sinusoidal activation functions of the form
\(
\sigma(x) = \sin(\omega_0 x),
\)
with a frequency scaling factor $\omega_0 = 20$, enabling accurate representation of high-frequency spatiotemporal features in the solution fields. The network size was selected to ensure sufficient expressive capacity for approximating a wide variety of input–output function mappings. No evidence of overfitting was observed for the considered training regime, as indicated by the consistent behavior of training and validation losses (see Fig.~\ref{fig:train_val_losses}).

This design choice is motivated by the structure of the DeepONet approximation, in which the trunk network can be interpreted as generating a set of coordinate-dependent basis functions, while the branch network produces the corresponding expansion coefficients conditioned on the input function. The use of sinusoidal activations in the trunk, therefore, enhances the representational capacity of the learned basis toward oscillatory and sharp spatiotemporal features \cite{Sitzmann2020ImplicitFunctions}. To ensure stable training of the sinusoidal network, SIREN weight initialization is adopted for the trunk network, while the branch network employs Kaiming initialization consistent with the SiLU activation.

The final DeepONet output is passed through a bounded nonlinear activation function (sigmoid) to ensure physically admissible predictions, constrained between 0 and 1.

\begin{table}[h!]
\centering
\caption{Architecture specifications of the DeepONet}
\label{tab:architecture}
\begin{tabular}{lcc}
\toprule
\textbf{Component} & \textbf{Feature} & \textbf{Value} \\
\midrule
\textbf{Trunk network} 
    & Hidden layers & 6 \\
    & Neurons per layer & 200 \\
    & Latent dimension & 100 \\
    & Activation function & Sine ($\omega_0 = 20$) \\
    & Weight initialization & SIREN \\
\midrule
\textbf{Branch network} 
    & Hidden layers & 6 \\
    & Neurons per layer & 200 \\
    & Latent dimension & 100 \\
    & Activation function & SiLU \\
    & Weight initialization & Kaiming \\
\bottomrule
\end{tabular}
\end{table}

\subsubsection{Training Procedure}
\label{sec:training procedure}
Training was performed in a supervised and data-driven mode by minimizing a weighted sum of (i) an initial-condition consistency loss and (ii) a full-field data loss computed against the reference solutions.

The initial condition loss enforces that the predicted gas-phase field at $\tau^*=0$ matches the input initial condition, with MSE indicating the mean squared error:
\begin{equation}
L_{\mathrm{ic}}
=
\mathrm{MSE}\!\left(
\widehat{\mathcal{G}}_g\!\left(C_g^{*0}\right)(\xi^*,0),
\;
C_g^{*0}(\xi^*)
\right),
\end{equation}

The data loss penalizes the discrepancy between the predicted and reference solution fields over the full spatiotemporal domain:
\begin{equation}
L_{\mathrm{data}}
=
\mathrm{MSE}\!\left(
\widehat{\mathcal{G}}_g\!\left(C_g^{*0}\right)(\xi^*,\tau^*),
\;
C_g^{*}(\xi^*,\tau^*)
\right).
\end{equation}

The total loss minimized during training is
\begin{equation}
L_{\mathrm{tot}}=\lambda_{\mathrm{ic}}\,L_{\mathrm{ic}}+\lambda_{\mathrm{data}}\,L_{\mathrm{data}},
\end{equation}
The specific values of the weights were selected based on an ablation study in which different combinations of weights were tested. The best performances were achieved for $\lambda_{\mathrm{ic}}=3$ and $\lambda_{\mathrm{data}}=1$.

Optimization was carried out using the Adam optimizer with a learning rate $10^{-4}$ and no weight decay. An adaptive learning rate scheduler was employed to reduce the step size when the loss plateaued, down to a minimum learning rate of $10^{-7}$. 
Training was run for a maximum of $5\times 10^{5}$ epochs with early stopping based on the validation loss. The validation loss was evaluated every $100$ epochs using the same loss definitions, and the model parameters corresponding to the lowest validation loss were saved. Training was terminated when no improvement in validation loss was observed for $10^{4}$ consecutive epochs.

The entire work was conducted in Python (version 3.12.9), using PyTorch (version 2.6.0+cu126) \cite{Paszke2019PyTorch:Library} as the primary deep learning framework. All simulations and training were performed on a workstation equipped with an NVIDIA GeForce RTX 4060 Ti GPU (NVIDIA Corporation, Santa Clara, CA, USA) (16 GB memory, CUDA version 12.4, driver version 550.142).

\section{Results}\label{sec:results}
The performance of the DeepONet is evaluated on both the test and the expanded-OOD datasets described in Sec.~\ref{subsec:data_generation}. Parity plots are used for straightforward qualitative visual information, while the mean $L^2$ relative error (see \autoref{sec:appendices}) is employed to provide a quantitative global metric relative to the specific dataset case. Given the tight coupling between the solid and gas phase, resulting in very similar dynamics, presenting results for both operators would be redundant. Therefore, only the gas phase operator results will be presented.
 
\begin{figure}[ht!]
\centering
\includegraphics[width=0.99\textwidth]{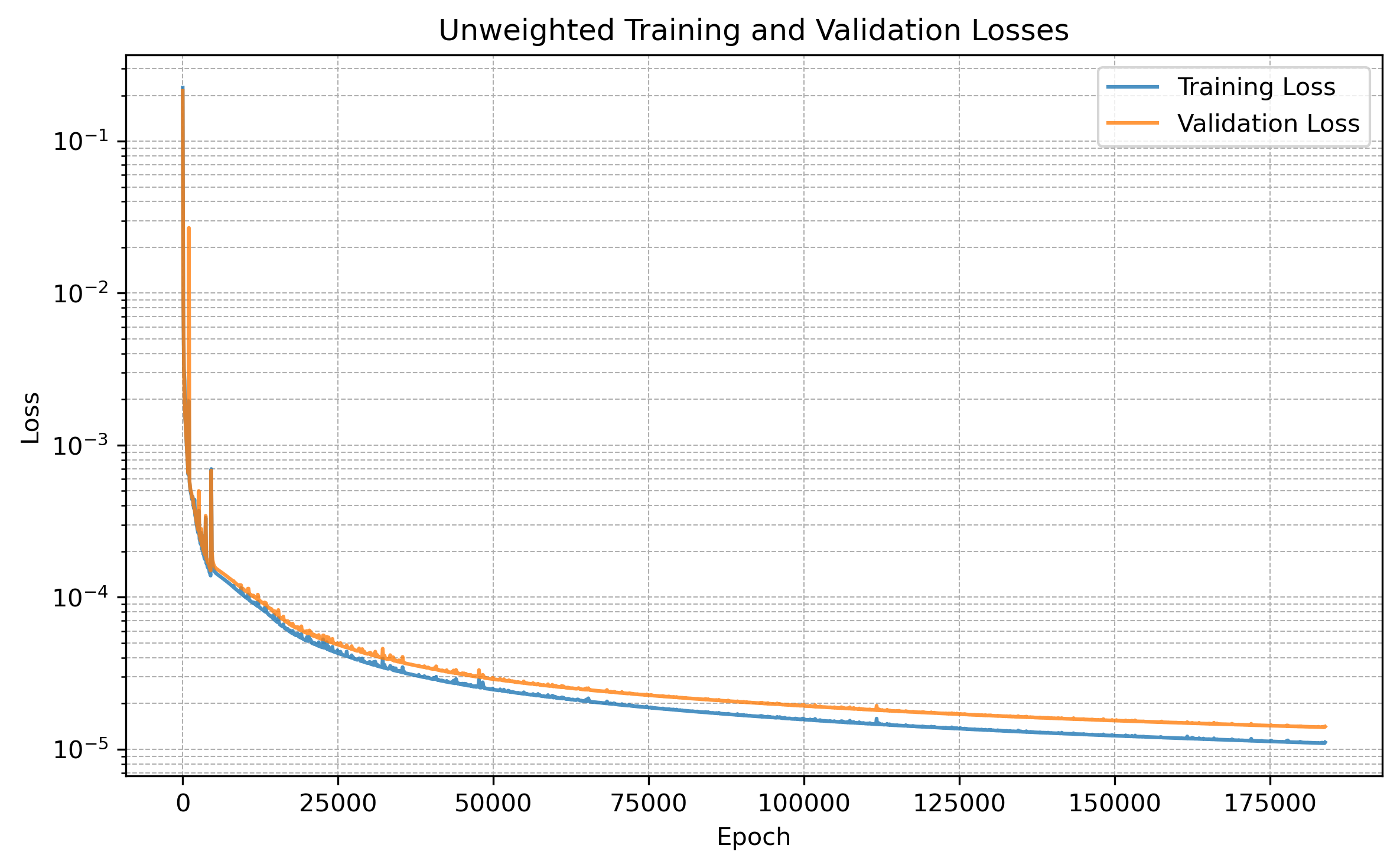}
\caption{Training and validation losses trends per epoch for the gas phase DeepONet.}\label{fig:train_val_losses}
\end{figure}

The training and validation losses as functions of the training epoch are shown in Figure~\ref{fig:train_val_losses}, while detailed information on the training time and final loss values is reported in Table~\ref{tab:trainingspec}. 
Both residuals show a sharp decrease during the first 10,000 epochs with some perturbations. Thereafter, a constant decline is observed, without signs of overfitting \cite{Goodfellow2016DeepLearning}. From epoch 150,000 to epoch 175,000, the rate of decrease is only $1.01\times10^{-5}$ every 10,000 epochs. Due to the low trade-off between the loss improvements and the training time, we opted for a manual early stopping strategy after 183,900 epochs.

\begin{table}[h!]
\centering
\caption{Training specifications for the DeepONet relative to the gas phase}
\label{tab:trainingspec}
\begin{tabular}{lcc}
\toprule
\textbf{Trainer} & \textbf{Specification} & \textbf{Value} \\
\midrule
\textbf{ADAM} 
    & Simulation time & 98 min 5 s \\
    & Smallest training loss & 2.77e-05 \\
    & Smallest validation loss & 1.39e-05 \\
    & Smallest data loss & 2.59e-06 \\
    & Smallest initial condition loss & 8.37e-06 \\
    & Epoch & 183{,}900 \\
\bottomrule
\end{tabular}
\end{table}

\subsection{Test Dataset Results}
\begin{figure}[H]
\centering
\includegraphics[width=0.7\textwidth]{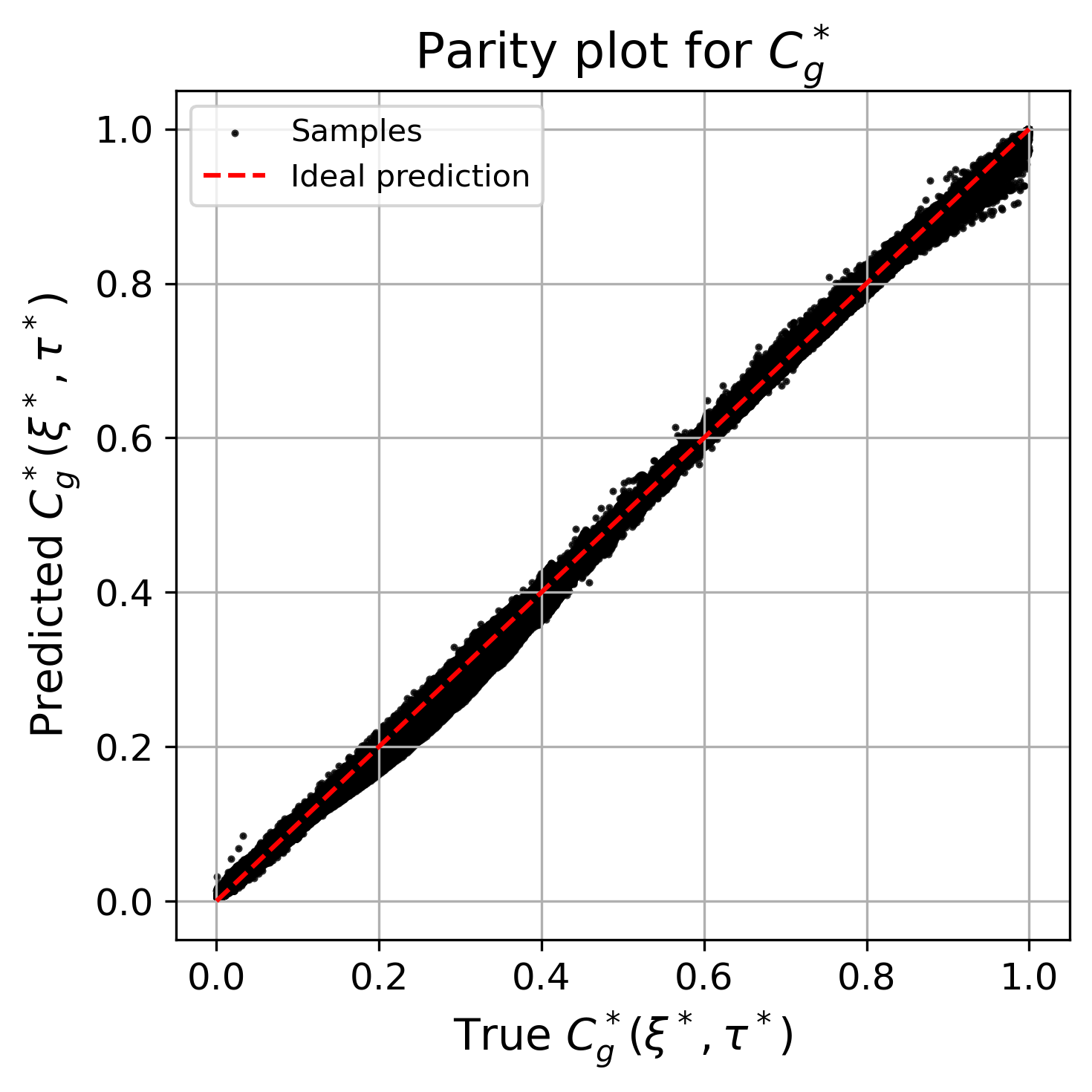}
\caption{Parity plot of $C_g^*$ for the test set, with an average $L^2$ relative error of $0.1684\%$. }\label{fig:test_parity}
\end{figure}

The model demonstrates accurate and consistent predictive performance throughout the test dataset. As shown in Fig.~\ref{fig:test_parity}, the parity plot reveals a strong agreement between predicted and ground-truth values, with the majority of samples closely clustered along the bisector and no evident outliers or systematic bias. This qualitative evidence is supported by a low $L^2$ relative error of $0.1684\%$ over the test set, providing quantitative confirmation of the model accuracy and its robust generalization performance.

\begin{figure}[H]
\centering
\begin{subfigure}[t]{1\textwidth}
    \centering
    \includegraphics[width=\linewidth]{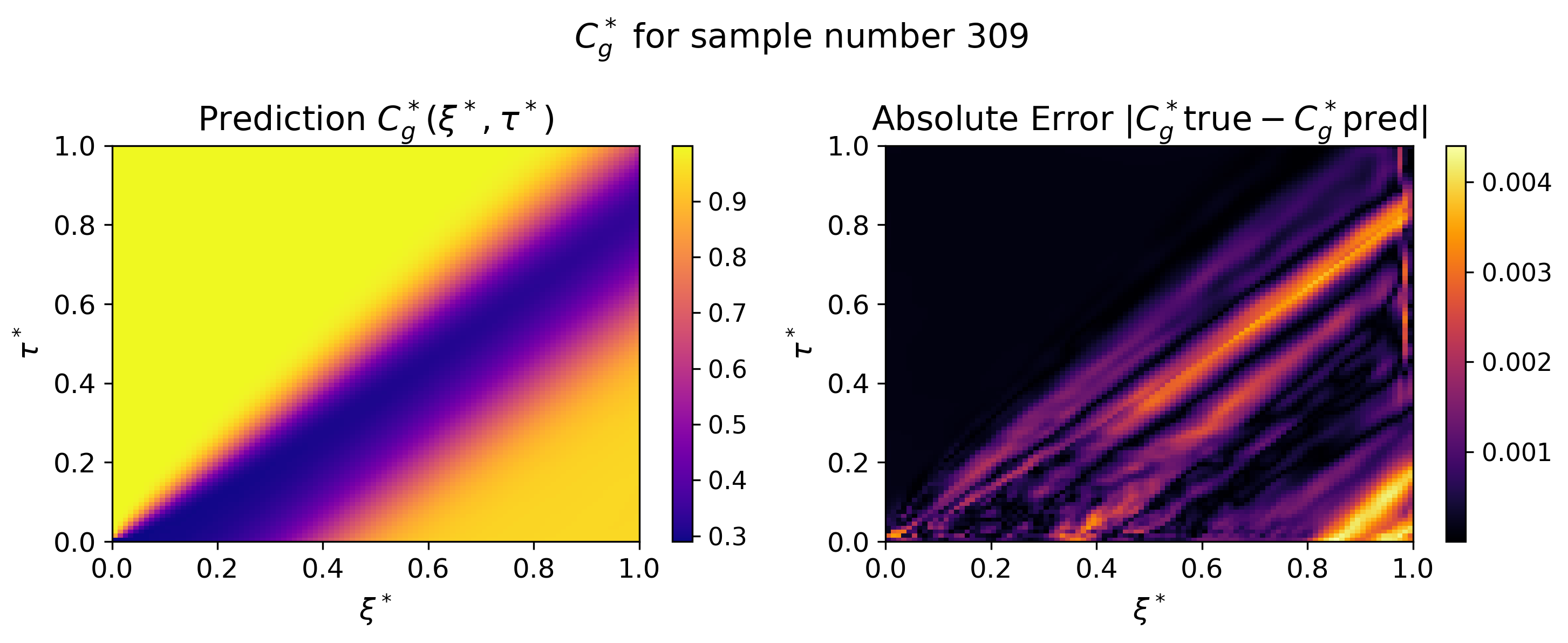}
    \caption{}
    \label{fig:heatmaps309}
\end{subfigure}
\hfill
\begin{subfigure}[t]{0.7\textwidth}
    \centering
    \includegraphics[width=\linewidth]{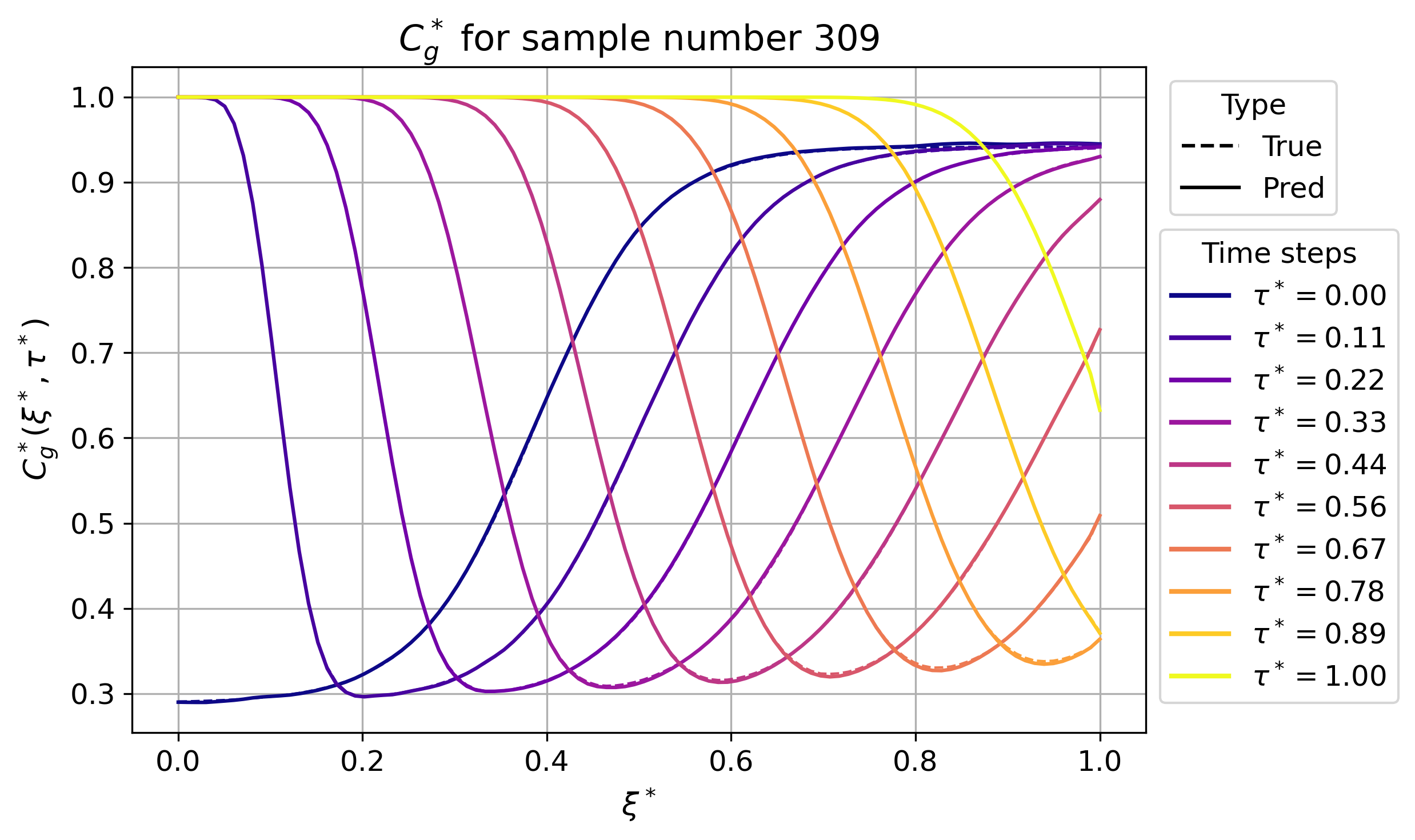}
    \caption{}
    \label{fig:snapshots309}
\end{subfigure}
\caption{Heatmaps (\ref{fig:heatmaps309}) of the prediction and the absolute point-wise error, and time snapshots (\ref{fig:snapshots309}) of the true and predicted results for a test example with a sigmoidal initial condition function.}
\label{fig:snapheat309}
\end{figure}

When evaluated on monotonic initial condition functions, as illustrated by the representative example in Fig.~\ref{fig:snapheat309}, the model demonstrates strong predictive performance overall. The absolute point-wise error heatmap in the right graph of Fig.~\ref{fig:heatmaps309} shows negligible errors in the region above the space–time bisector, where the gas-phase concentration remains approximately constant and the solution is smooth.

In contrast, the region below the bisector corresponds to the propagation of the adsorption front, where sharp concentration gradients and evolving tailing effects dominate the dynamics. This region is inherently more challenging to learn due to the strong spatiotemporal variations and non-smooth features of the solution. Accordingly, higher errors are observed in this area; however, the maximum absolute point-wise error remains below $4.5\times10^{-3}$. Notably, the error is not randomly distributed but appears localized along distinct space–time bands. This structured error pattern indicates that the model accurately captures the bulk transport behavior, while residual discrepancies are confined to the most dynamically complex regions of the adsorption front.

\begin{figure}[H]
\centering
\begin{subfigure}[t]{1\textwidth}
    \centering
    \includegraphics[width=\linewidth]{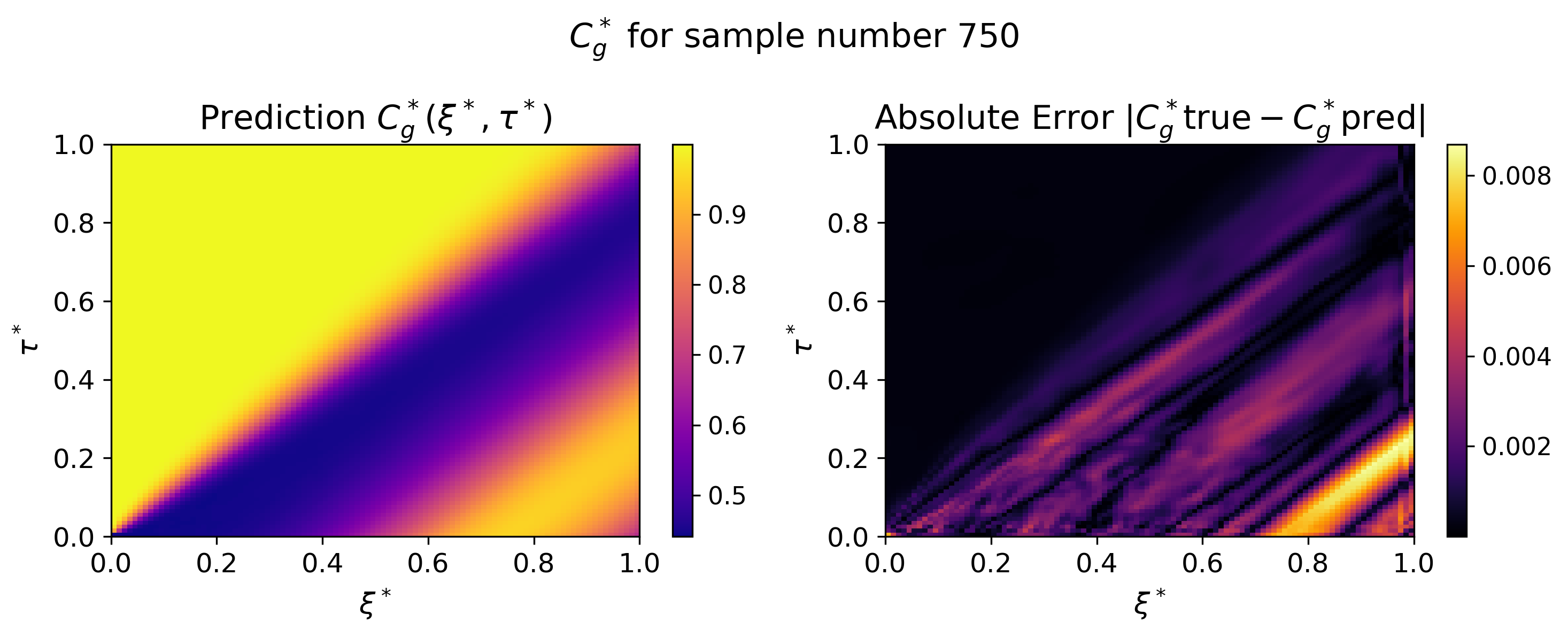}
    \caption{}
    \label{fig:heatmaps750}
\end{subfigure}
\hfill
\begin{subfigure}[t]{0.7\textwidth}
    \centering
    \includegraphics[width=\linewidth]{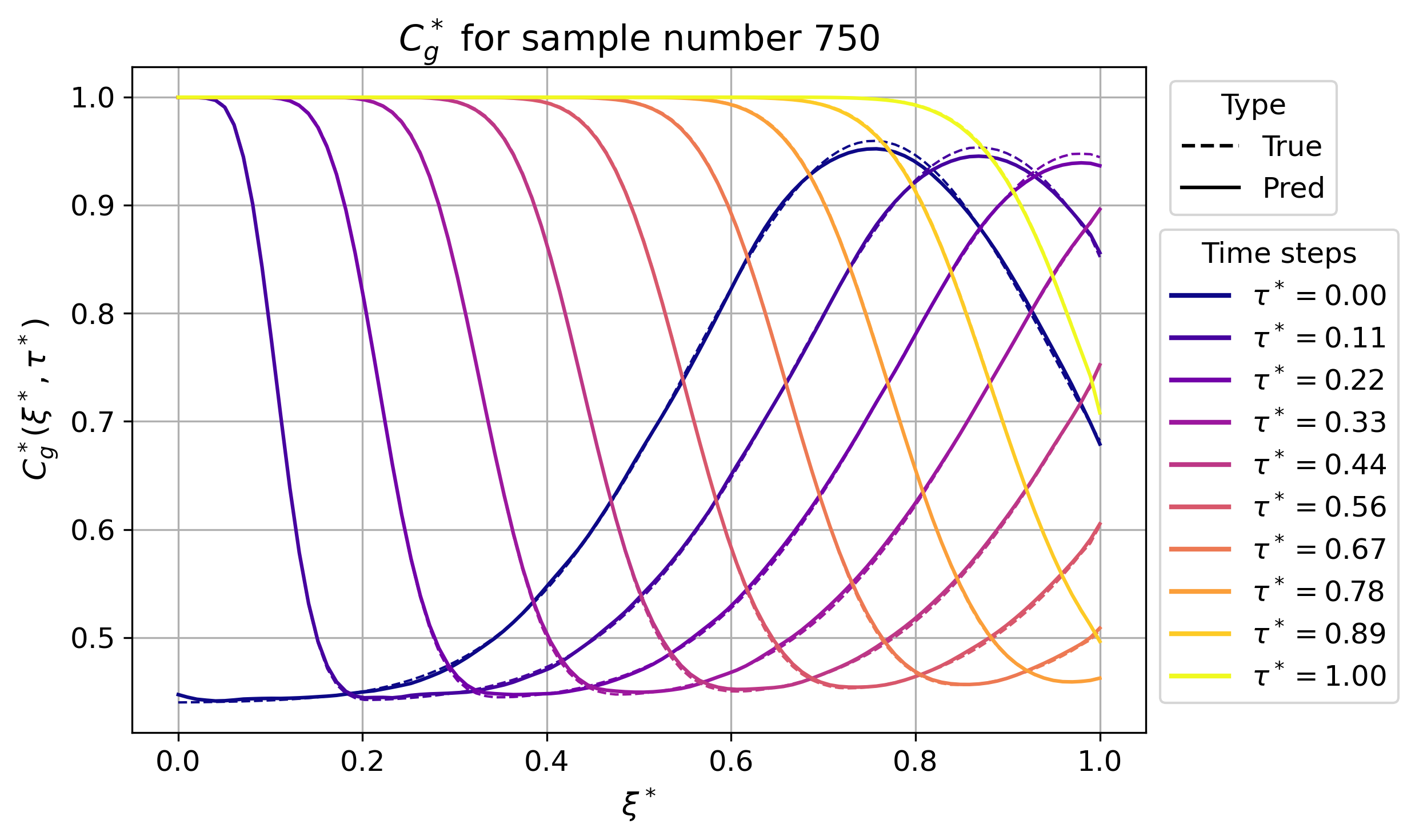}
    \caption{}
    \label{fig:snapshots750}
\end{subfigure}
\caption{Heatmaps (\ref{fig:heatmaps750}) of the prediction and the absolute pointwise error, and time snapshots (\ref{fig:snapshots750}) of the true and predicted results for a test example with a Gaussian-like initial condition function.}
\label{fig:snapheat750}
\end{figure}

The model also demonstrates strong performance for non-monotonic inputs, as illustrated by the representative example shown in Fig.~\ref{fig:snapheat750}. In this case, the overall error distribution follows a trend comparable to that observed in the monotonic example. However, regions corresponding to the peaks of the Gaussian-shaped prediction domain exhibit higher errors, with maximum absolute point-wise values exceeding $8\times10^{-3}$.
The higher error in the vicinity of the Gaussian peak indicates a localized limitation in accurately reproducing sharp spatial features. Such behavior is consistent with the well-known spectral bias of neural networks, which tend to preferentially learn low-frequency components of the target function, while higher-frequency content is captured less accurately \cite{Xu2025OverviewLearning}. 

\subsection{Extended Dataset Results}
The model achieved strong predictive performance also when evaluated on the extended out-of-distribution (OOD) dataset, with a average $L^2$ relative error of $2.282 \%$. Remarkably, the model demonstrated the ability to generalize to input functions that were not encountered during either training or validation.

\begin{figure}[H]
\centering
\includegraphics[width=0.7\textwidth]{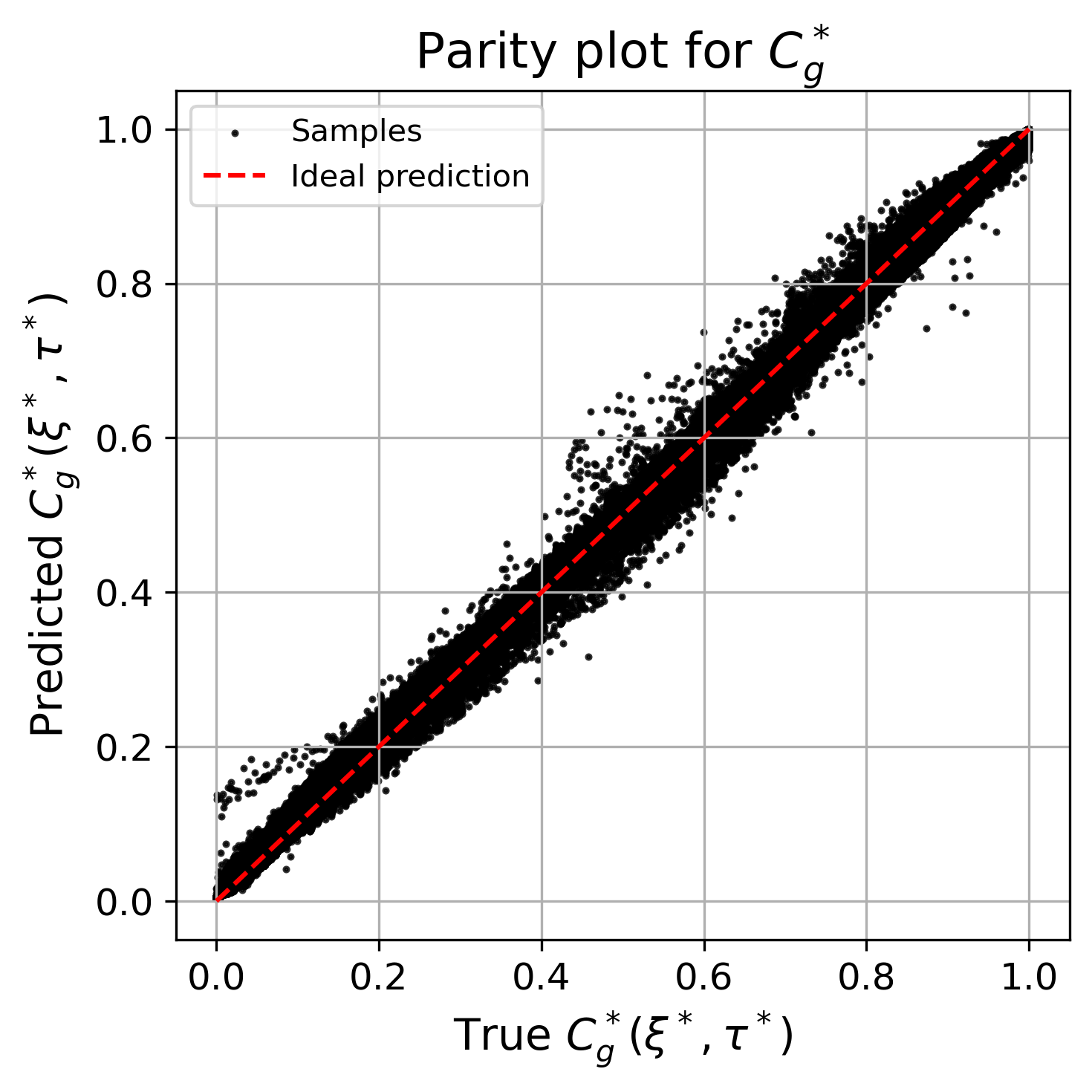}
\caption{Parity plot of $C_g^*$ for the extended dataset, with an average $L^2$ relative error of $0.4212\%$.
}\label{fig:extended_parity}
\end{figure}

The parity plot corresponding to the extended dataset in Fig.~\ref{fig:extended_parity} displays an increased dispersion of the predicted values around the bisector relative to the test-dataset parity plot shown in Fig \ref{fig:test_parity}. Nevertheless, no systematic bias can be identified. This behavior is attributed to the inclusion of entirely unseen initial condition samples in the extended dataset, which naturally increases the variability of the prediction error while preserving overall accuracy.

\begin{figure}[H]
\centering
\begin{subfigure}[t]{1\textwidth}
    \centering
    \includegraphics[width=\linewidth]{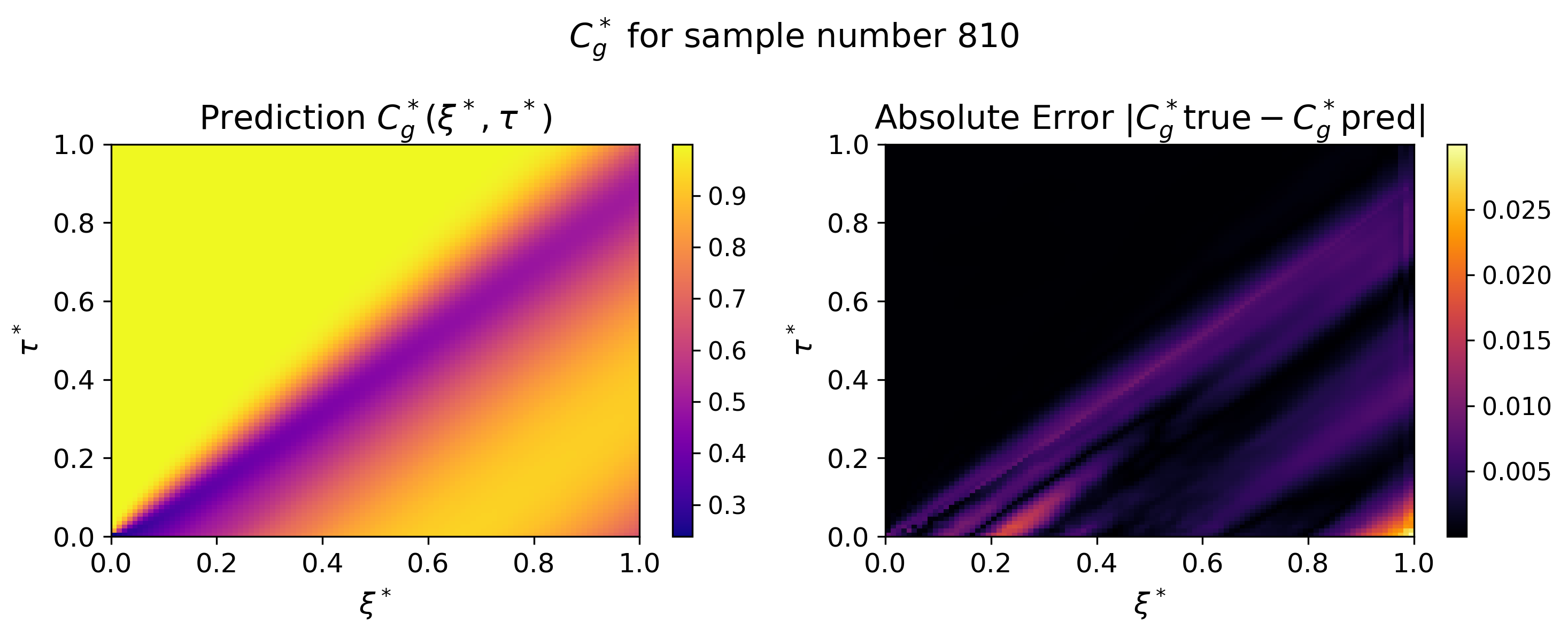}
    \caption{Heatmaps}
    \label{fig:heatmaps810}
\end{subfigure}
\hfill
\begin{subfigure}[t]{0.7\textwidth}
    \centering
    \includegraphics[width=\linewidth]{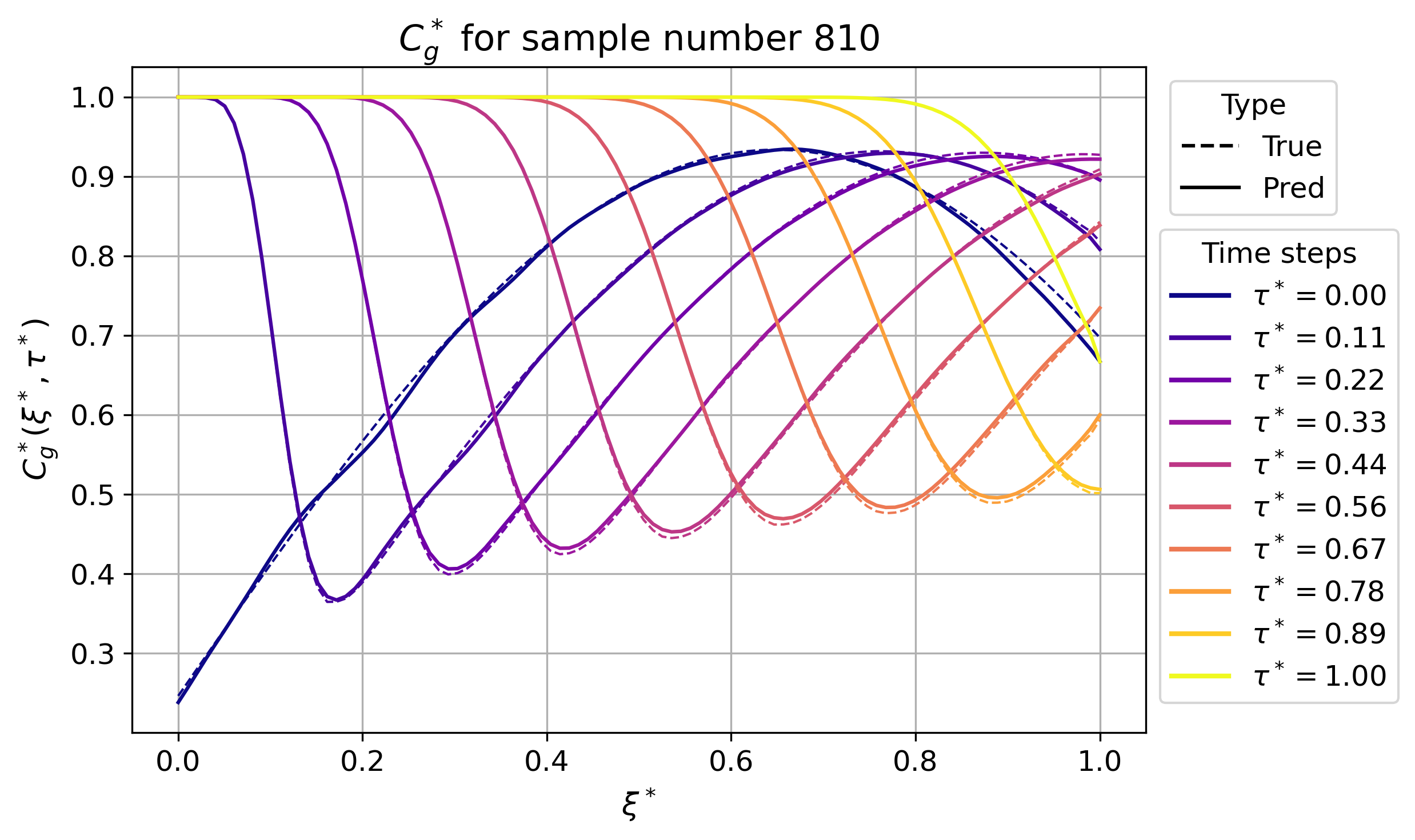}
    \caption{Snapshots}
    \label{fig:snapshots810}
\end{subfigure}
\caption{Heatmaps (\ref{fig:heatmaps810}) of the prediction and the absolute point-wise error, and time snapshots (\ref{fig:snapshots810}) of the true and predicted results for a test example with a sinusoidal initial condition function.}
\label{fig:snapheatsine810}
\end{figure}

Fig.~\ref{fig:snapheatsine810} shows a representative reconstruction of an unseen sine function. The relative $L^2$ error for this case amounts to $0.4212 \%$, and the overall spatial distribution of the error, as observed in the absolute error heatmap, remains consistent with the patterns identified in the previous cases.

\begin{figure}[H]
\centering
\begin{subfigure}[t]{1\textwidth}
    \centering
    \includegraphics[width=\linewidth]{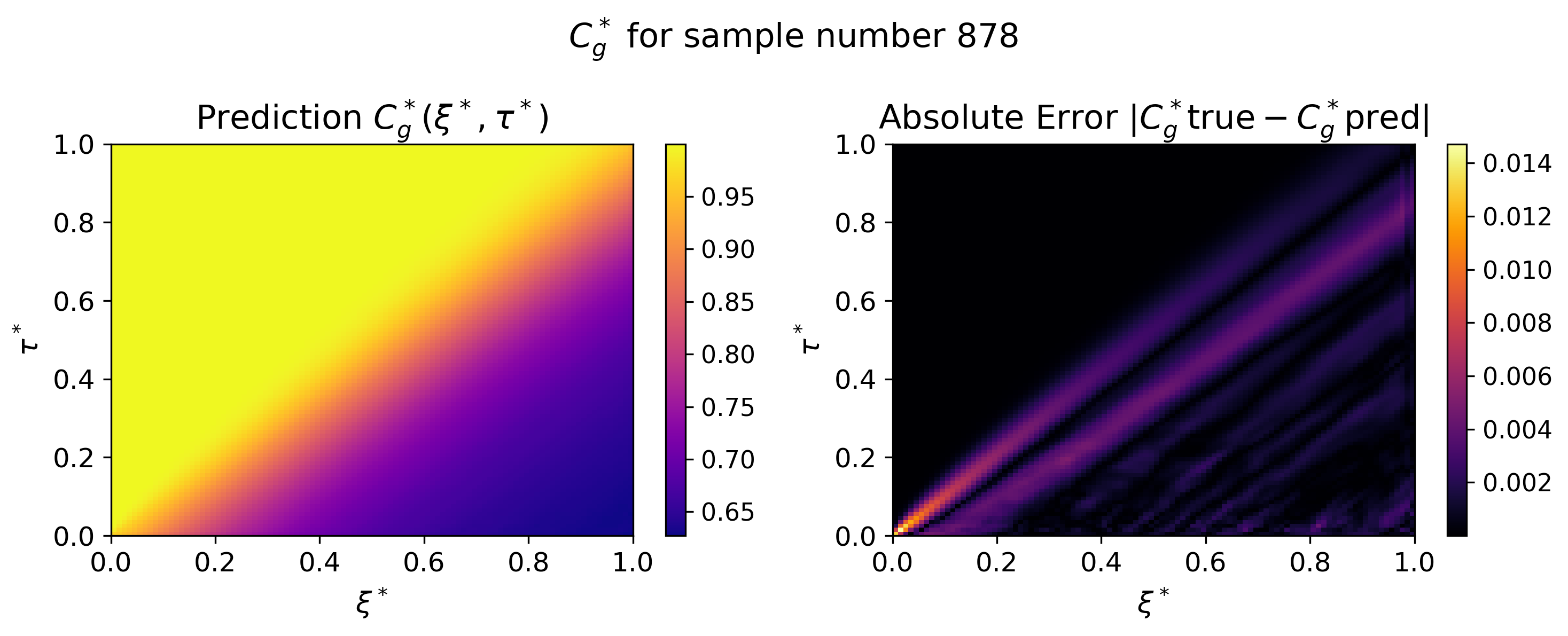}
    \caption{Heatmaps}
    \label{fig:heatmaps878}
\end{subfigure}
\hfill
\begin{subfigure}[t]{0.7\textwidth}
    \centering
    \includegraphics[width=\linewidth]{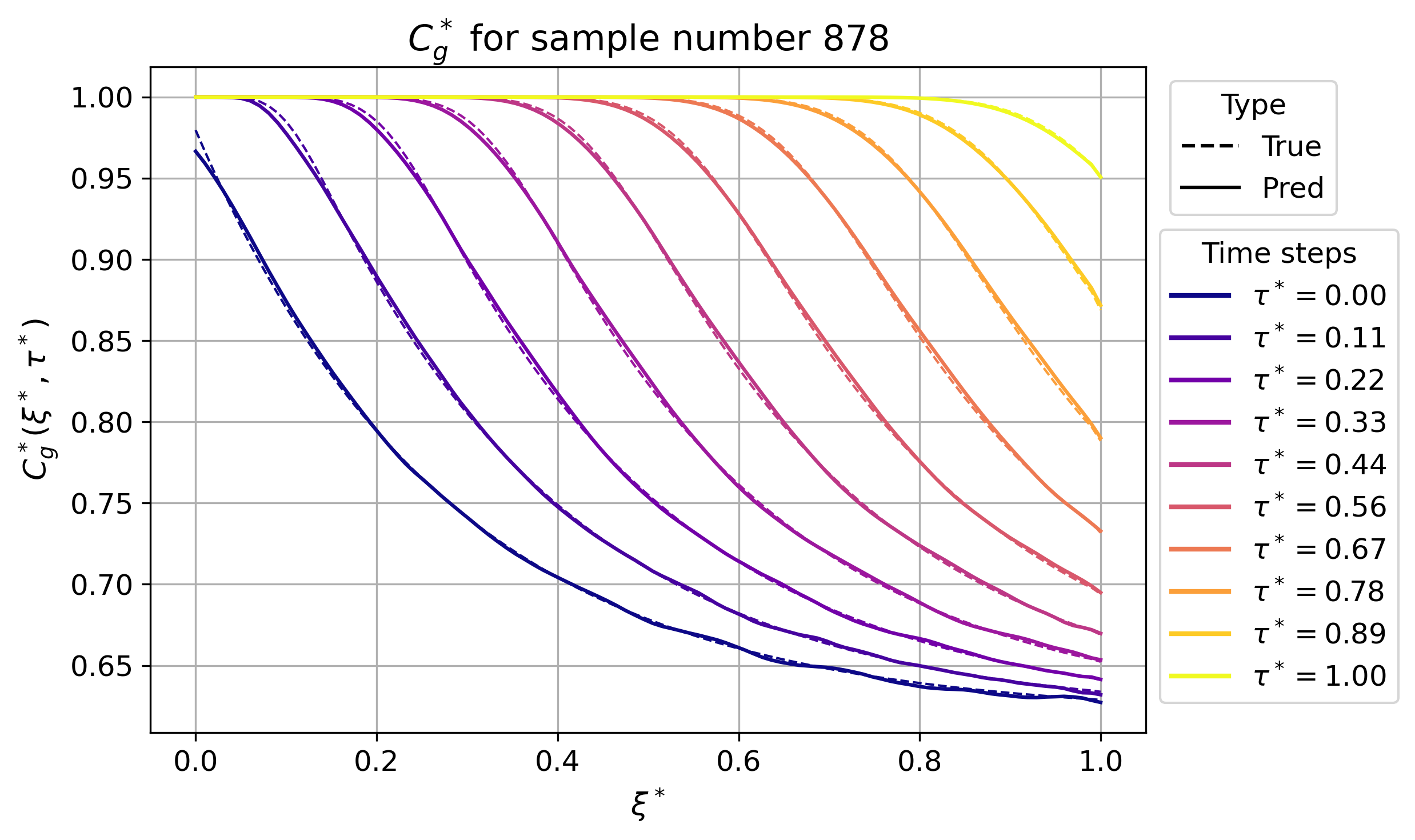}
    \caption{Snapshots}
    \label{fig:snapshots878}
\end{subfigure}
\caption{Heatmaps (\ref{fig:heatmaps878}) of the prediction and the absolute point-wise error, and time snapshots (\ref{fig:snapshots878}) of the true and predicted results for a test example with an exponential initial condition function.}
\label{fig:snapheat878}
\end{figure}

Moreover, the network achieves good predictive performance on unseen functions that belong to initial condition functional classes already represented in the training set. In particular, the example reported in Fig.~\ref{fig:snapheat878} yields an $L^2$ relative error of $0.1802\%$.

\section{Discussion and conclusions}\label{sec:discussion}
In this work, we demonstrated that Deep Operator Networks can successfully learn solution operators for adsorption-driven PDE systems when conditioned on a wide range of physically admissible initial conditions. By formulating the transient adsorption problem as an operator mapping from initial concentration profiles to spatiotemporal solution fields, DeepONets were shown to provide accurate reconstructions of both gas and solid phase dynamics. More importantly, the trained models exhibited robust functional generalization beyond the training distribution, including for initial condition functions characterized by functional forms not encountered during training.

Despite these advantages, the present study also highlights limitations that may warrant further investigation. The largest prediction errors were observed in a limited number of specific cases, which provide insight into the current boundaries of the proposed DeepONet surrogate.

\begin{figure}[H]
\centering
\includegraphics[width=1\textwidth]{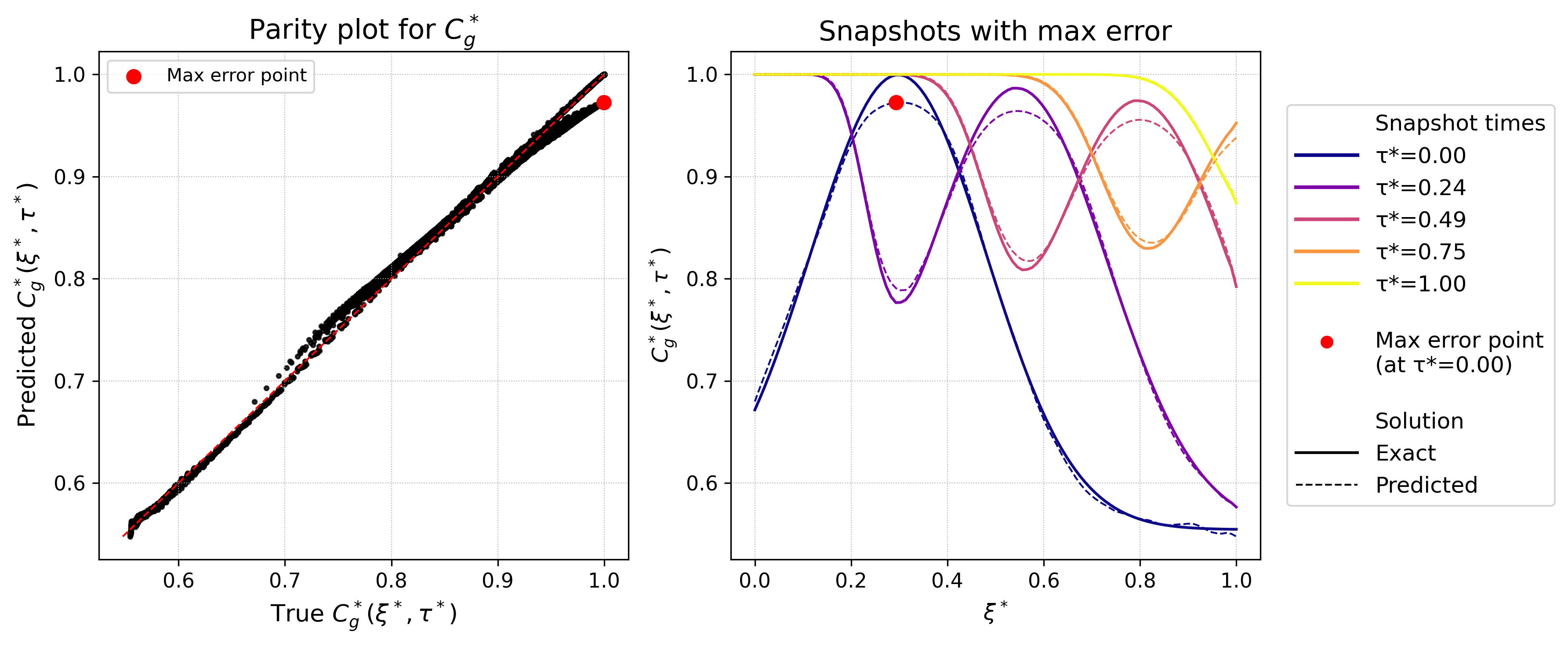}
\caption{Parity plot and time snapshots with highlighted maximum absolute point-wise error point for a representative example of the test dataset.}\label{fig:worst_case_1}
\end{figure}

Figure \ref{fig:worst_case_1} reports a representative worst-case prediction within the test dataset. Although the global accuracy remains high, the maximum error is observed at the initial time (\( \tau^*=0\)) and is associated with initial conditions exhibiting sharp spatial variations. The corresponding snapshots reveal localized discrepancies near high-curvature regions of the concentration profile, while the subsequent temporal evolution is accurately captured. This behavior indicates that the dominant source of error originates from the reconstruction of high-frequency spatial features in the initial condition, rather than from the learned transient dynamics. Such localized inaccuracies are consistent with known challenges in operator learning and neural surrogates, where spectral bias can hinder the representation of high-frequency components and sharp localized features, leading to good global metrics but reduced local fidelity near steep gradients or narrow peaks \cite{Rahaman2018OnNetworks}. Similar smoothing effects have also been reported for neural operators in hyperbolic settings with sharp fronts/discontinuities \cite{Chauhan2025NeuralStudy}. These observations suggest that the dominant error mechanisms are primarily linked to the representation of high-frequency spatial features in the input space rather than to deficiencies in the learned dynamics. 

\begin{figure}[H]
\centering
\includegraphics[width=1\textwidth]{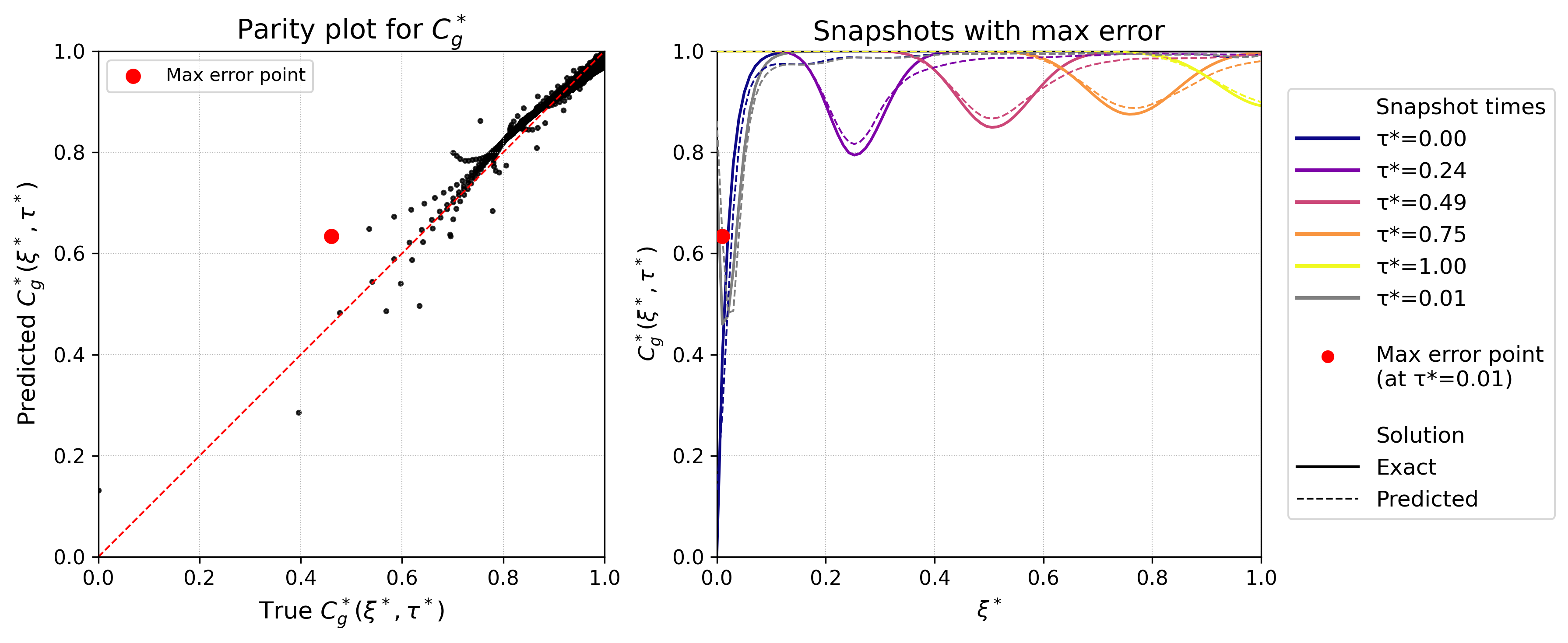}
\caption{Parity plot and time snapshots with highlighted maximum absolute point-wise error point for a representative example of the extended OOD dataset.}\label{fig:worst_case_2}
\end{figure}

A different error mechanism is observed in another worst-case extrapolation examples shown in Figure \ref{fig:worst_case_2}. In these cases, the gas-phase initial condition exhibits an extremely steep spatial gradient in close proximity to the inlet, while the boundary condition enforces a fixed Dirichlet value (\( \tau^*=0\)).  Due to the convective nature of the governing equations, this localized discrepancy is subsequently transported downstream, resulting in a visible propagation of the error across the spatiotemporal domain. Nevertheless, the magnitude of the error does not grow unbounded, nor does it trigger numerical instability, indicating that the learned operator remains dynamically stable even under severe out-of-distribution conditions.

Potential mitigation strategies have been reported in the literature and constitute promising directions for future work. For instance, the observed downstream propagation of the initial discrepancy is consistent with advection-dominated dynamics and could be alleviated by enforcing boundary conditions as hard constraints, thereby reducing mismatch at the inlet \cite{Brecht2023ImprovingConstraints}. Additionally, high-frequency reconstruction errors could potentially be mitigated by enriching the input representation of the branch network, for instance, through Fourier feature mappings or multi-resolution encodings, which have been shown to alleviate spectral bias in neural approximators \cite{Tancik2020FourierDomains}. Finally, using structured or reduced-order bases in the trunk network (e.g., POD-informed DeepONets) could further improve the representation of localized spatial features while preserving global accuracy \cite{Lu2022AData}.

Overall, Deep Operator Networks emerge as a promising and computationally efficient surrogate modeling framework for adsorption-based processes, particularly in the context of cyclic operation. By explicitly learning solution operators conditioned on physically admissible initial conditions, the proposed approach is well aligned with the sequential structure of adsorption cycles, where each step inherits its initial state from the preceding one.

From a computational perspective, purely data-driven DeepONets require the generation of a large number of high-fidelity numerical solutions during the training phase, which represents their primary computational cost. However, this offline cost can be amortized in scenarios where repeated evaluations of the governing PDEs are required, such as cyclic steady-state simulations, parametric studies, and process optimization. Once trained, the inference speed of DeepONets is orders of magnitude faster than that of conventional PDE solvers, enabling rapid exploration of the solution space and efficient propagation of state variables across cycle steps.

\section{Declarations}
\subsection{Availability of data and materials}
The datasets used and/or analyzed during the current study are available from the corresponding author upon reasonable request.
\subsection{Competing interests}
The authors declare that they have no competing interests.
\subsection{Funding}
We thank Shell Global Solutions International BV for financial support.
\subsection{Authors' contributions}
MG was responsible for project management, methodological development, code implementation, data analysis, and drafted the original manuscript.
BC contributed to model development, code implementation, numerical testing, data analysis, and manuscript writing.
IR and MvSA provided scientific supervision, critically reviewed the manuscript, and contributed to the interpretation of the results.
All authors read and approved the final manuscript.
\subsection{Acknowledgments}
Not applicable.

\section*{List of Abbreviations}
\begin{longtable}{ll}
\hline
Abbreviation & Description \\
\hline
PDE  & Partial Differential Equation \\
TSA  & Temperature Swing Adsorption \\
VSA  & Vacuum Swing Adsorption \\
TVSA & Temperature Vacuum Swing Adsorption \\
CSS  & Cyclic Steady State \\
OOD  & Out-of-Distribution \\
MSE  & Mean Squared Error \\
RMSE & Root Mean Squared Error \\
LDF  & Linear Driving Force \\
\hline
\end{longtable}

\section*{List of Symbols}
\begin{longtable}{ll}
\hline
Symbol & Description \\
\hline
$C_g$ & Gas-phase molar concentration [$\mathrm{mol\,m^{-3}}$] \\
$C_s$ & Solid-phase molar concentration per unit adsorbent volume [$\mathrm{mol\,m^{-3}}$] \\
$C_s^R$ & Adsorbed-phase concentration at the external particle surface [$\mathrm{mol\,m^{-3}}$] \\
$C_0$ & Reference concentration [$\mathrm{mol\,m^{-3}}$] \\
$C_g^*$ & Dimensionless gas-phase concentration \\
$C_s^*$ & Dimensionless solid-phase concentration \\
$C_{g}^{*0}(\xi^*)$ & Dimensionless gas-phase initial condition \\
$C_{s}^{*0}(\xi^*)$ & Dimensionless solid-phase initial condition \\
$t$ & Time [$\mathrm{s}$] \\
$x$ & Axial spatial coordinate [$\mathrm{m}$] \\
$\tau$ & Dimensionless time coordinate \\
$\xi$ & Dimensionless spatial coordinate \\
$\tau^*$ & Normalized dimensionless time \\
$\xi^*$ & Normalized dimensionless spatial coordinate \\
$\tau_0$ & Reference dimensionless time scaling \\
$\xi_0$ & Reference dimensionless spatial scaling \\
$v_x$ & Superficial gas velocity [$\mathrm{m\,s^{-1}}$] \\
$k_{\mathrm{gas}}$ & Gas–solid mass transfer coefficient [$\mathrm{m\,s^{-1}}$] \\
$\varepsilon_B$ & Packed bed porosity [-] \\
$K_{\mathrm{eq}}$ & Adsorption equilibrium constant [-] \\
$L$ & Length of the packed bed [$\mathrm{m}$] \\
$d_p$ & Particle diameter [$\mathrm{m}$] \\
$a_s$ & Specific particle surface area [$\mathrm{m^{-1}}$] \\
$t_{\mathrm{tot}}$ & Total simulation time [$\mathrm{s}$] \\
$f(x)$ & Analytical function\\
$g(x)$ & Analytical function after rescaling \\
$\mathcal{G}_g$ & Gas-phase solution operator \\
$\mathcal{G}_s$ & Solid-phase solution operator \\
$\widehat{\mathcal{G}}_g$ & DeepONet approximation of the gas-phase operator \\
$\widehat{\mathcal{G}}_s$ & DeepONet approximation of the solid-phase operator \\
$\phi_k(\xi^*,\tau^*)$ & Trunk network basis functions \\
$a_k(C_g^{*0})$ & Branch network coefficients \\
$\sigma(x)$ & Activation function\\
$p$ & Latent space dimension (number of basis functions) \\
$L_{\mathrm{ic}}$ & Initial condition loss \\
$L_{\mathrm{data}}$ & Data loss \\
$L_{\mathrm{tot}}$ & Total training loss \\
$\lambda_{\mathrm{ic}}$ & Weight of the initial condition loss \\
$\lambda_{\mathrm{data}}$ & Weight of the data loss \\
$N$ & Number of samples in a dataset \\
$M_x$ & Number of spatial grid points \\
$M_t$ & Number of temporal grid points \\
 $\overline r_g$ & Mean relative $L^2$ error for the gas phase \\
$\overline r_s$ & Mean relative $L^2$ error for the solid phase \\
\hline
\end{longtable}

\begin{appendices}
\section{Performance metrics}\label{sec:appendices}
The performance of the Deep Operator Networks is assessed using the
mean relative $L^2$ error, computed on discretized spatiotemporal solution
fields. All metrics are evaluated on uniform grids in space and time and
reported separately for the gas- and solid-phase operators.

We consider a test (or training) dataset consisting of $N$ examples, indexed
by $i=1,\dots,N$. For each example, the DeepONet approximates the solution
operators $\mathcal{G}_g$ and $\mathcal{G}_s$ to $\widehat{\mathcal{G}}_g$ and $\widehat{\mathcal{G}}_s$, mapping the initial condition
$\mathbf{C}^{*0,(i)}(\xi^*)$ to the corresponding gas- and solid-phase solution
fields. The solutions are evaluated on a uniform spatiotemporal grid defined
by $\{\xi_j^*\}_{j=1}^{M_x}$ and $\{\tau_k^*\}_{k=1}^{M_t}$.

For each example $i$, we define the discrete solution fields
\begin{align}
U^{(i)}_{g;j,k}
&:= \mathcal{G}_g\!\left(\mathbf{C}^{*0,(i)}\right)(\xi_j^*,\tau_k^*), \\
\widehat U^{(i)}_{g;j,k}
&:= \widehat{\mathcal{G}}_g\!\left(\mathbf{C}^{*0,(i)}\right)(\xi_j^*,\tau_k^*),
\end{align}
and analogously $U^{(i)}_{s;j,k}$ and $\widehat U^{(i)}_{s;j,k}$ for the solid
phase, where $\widehat U^{(i)}_{g;j,k}$ and $\widehat U^{(i)}_{s;j,k}$ represent the network's approximations. Each discrete field is represented as a matrix of size
$M_x \times M_t$.

The discrete $L^2$ norm of a matrix $A \in \mathbb{R}^{M_x \times M_t}$ is
defined as
\begin{equation}
\label{eq:L2_discrete}
\|A\|_2 :=
\left(
\sum_{j=1}^{M_x}
\sum_{k=1}^{M_t}
A_{j,k}^2
\right)^{1/2}.
\end{equation}

Using this definition, the relative $L^2$ error for example $i$ is computed
separately for the gas and solid phases as
\begin{align}
r^{(i)}_{g}
&:= \frac{\left\| \widehat U^{(i)}_{g} - U^{(i)}_{g} \right\|_2}
{\left\| U^{(i)}_{g} \right\|_2}, \\
r^{(i)}_{s}
&:= \frac{\left\| \widehat U^{(i)}_{s} - U^{(i)}_{s} \right\|_2}
{\left\| U^{(i)}_{s} \right\|_2}.
\end{align}

The overall performance metric reported for a given dataset is the mean
relative $L^2$ error over all examples,
\begin{equation}
\label{eq:mean_relative_L2}
\overline r_{g} := \frac{1}{N} \sum_{i=1}^{N} r^{(i)}_{g},
\qquad
\overline r_{s} := \frac{1}{N} \sum_{i=1}^{N} r^{(i)}_{s}.
\end{equation}

This metric provides a normalized measure of the approximation accuracy of
the learned operators and enables a consistent comparison of DeepONet
performance across different initial conditions and datasets.
\end{appendices}

\bibliography{references.bib}
\end{document}